\pdfoutput=1

\documentclass[11pt]{article}

\usepackage{acl}

\usepackage{times}
\usepackage{latexsym}
\usepackage{hyperref}

\usepackage[T1]{fontenc}

\usepackage[utf8]{inputenc}
\usepackage{float}

\usepackage{microtype}
\usepackage{algorithm}
\usepackage{algorithmic}
\usepackage{amsmath}
\usepackage{tabularx}
\usepackage{subfigure}
\usepackage{arydshln}
\usepackage{multirow}

\usepackage{newfloat}
\usepackage{listings}

\usepackage{booktabs}
\usepackage{array}
\newcommand{\PreserveBackslash}[1]{\let\temp=\\#1\let\\=\temp}
\newcolumntype{C}[1]{>{\PreserveBackslash\centering}p{#1}}
\newcolumntype{L}[1]{>{\PreserveBackslash\raggedright}p{#1}}

\usepackage{graphicx} 
\usepackage{natbib}  
\usepackage{caption} 

\title{Sentence-aware Contrastive Learning for \\ Open-Domain Passage Retrieval}

\author{Bohong Wu\textsuperscript{1,2,3}, Zhuosheng Zhang\textsuperscript{1,2,3}, Jinyuan Wang\textsuperscript{1,2,3}, Hai Zhao\textsuperscript{1,2,3,\thanks{*Corresponding author. This paper was partially supported by Key Projects of National Natural Science Foundation of China (U1836222 and 61733011).}} \\
    	$^{1}$ Department of Computer Science and Engineering, Shanghai Jiao Tong University \\
    	$^{2}$ Key Laboratory of Shanghai Education Commission for Intelligent Interaction \\
and Cognitive Engineering, Shanghai Jiao Tong University \\
        $^{3}$ MoE Key Lab of Artificial Intelligence, AI Institute, Shanghai Jiao Tong University \\
	\texttt{chengzhipanpan@sjtu.edu.cn,zhangzs@sjtu.edu.cn} \\
	\texttt{steve\_wang@sjtu.edu.cn,zhaohai@cs.sjtu.edu.cn}
}

\begin{document}
\maketitle

\begin{abstract}
	Training dense passage representations via contrastive learning has been shown effective for Open-Domain Passage Retrieval (ODPR). Existing studies focus on further optimizing by improving negative sampling strategy or extra pretraining. However, these studies keep unknown in capturing passage with internal representation conflicts from improper modeling granularity. This work thus presents a refined model on the basis of a smaller granularity, contextual sentences, to alleviate the concerned conflicts. In detail, we introduce an in-passage negative sampling strategy to encourage a diverse generation of sentence representations within the same passage. Experiments on three benchmark datasets verify the efficacy of our method, especially on datasets where conflicts are severe. Extensive experiments further present good transferability of our method across datasets.
\end{abstract}

\section{Introduction}

Open-Domain Passage Retrieval (ODPR) has recently attracted the attention of researchers for its wide usage both academically and industrially \cite{lee2019latent, yang2017anserini}. Provided with an extremely large text corpus that composed of millions of passages, ODPR aims to retrieve a collection of the most relevant passages as the evidences of a given question. 

With recent success in pretrained language models (PrLMs) like BERT \cite{devlin2019bert}, RoBERTa \cite{liu2019roberta}, dense retrieval techniques have achieved significant better results than traditional lexical based methods, including TF-IDF \cite{ramos2003using} and BM25 \cite{robertson2009probabilistic}, which totally neglect semantic similarity. Thanks to the Bi-Encoder structure, dense methods \cite{lee2019latent, guu2020realm, karpukhin2020dense} encode the Wikipedia passages and questions separately, and retrieve evidence passages using similarity functions like the inner product or cosine similarity. Given that the representations of Wikipedia passages could be precomputed, the retrieval speed of dense approaches could be on par with lexical ones.




Previous approaches often pretrain the Bi-Encoders with a specially designed pretraining objective, Inverse Cloze Task (ICT) \cite{lee2019latent}. More recently, DPR \cite{karpukhin2020dense} adopts a simple but effective contrastive learning framework, achieving impressive performance without any pretraining. Concretely, for each question $q$, several positive passages $p^{+}$ and hard negative passages $p^{-}$ produced by BM25 are pre-extracted. By feeding the Bi-Encoder with $(q, p^{+}, p^{-})$ triples, DPR simultaneously maximizes the similarity between the representation of $q$ and corresponding $p^{+}$, and minimizes the similarity between the representations of $q$ and all $p^{-}$. Following such contrastive learning framework, many researchers are seeking further improvements for DPR from the perspective of sampling strategy \cite{xiong2020approximate, lu2020neural, tang2021improving, qu2021rocketqa} or extra pretraining \cite{sachan2021end}, or even using knowledge distillation \cite{izacard2021distilling, yang2021neural}.

\begin{figure}[tp]
	\centering
	\includegraphics[width=1.0\linewidth]{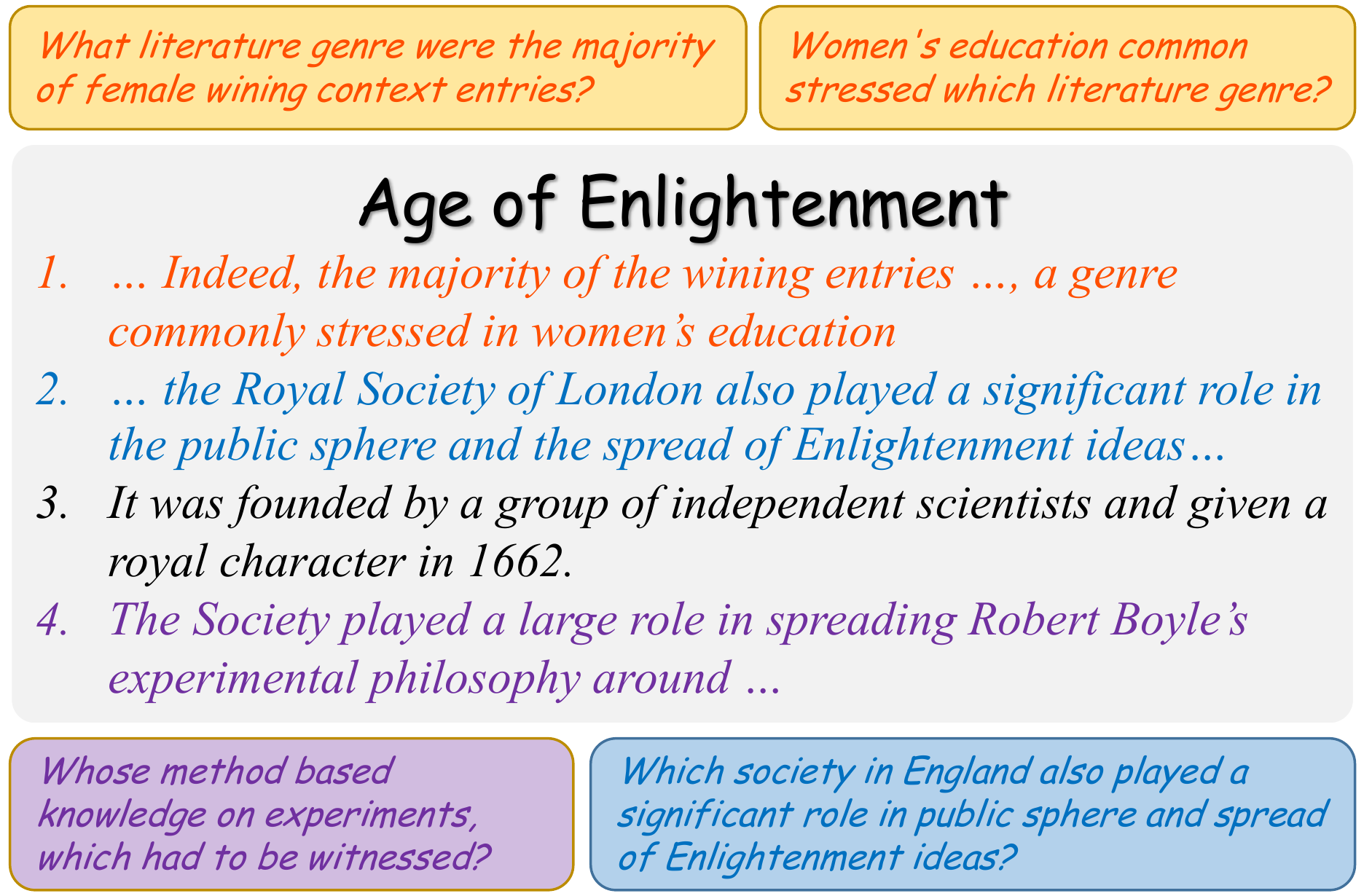}
	\caption{A sample from SQuAD. Different colors indicate the questions/sentences focus on different topics.}
	\label{fig:sample} 
\end{figure}

However, these studies fail to realize that there exist severe drawbacks in the current contrastive learning framework adopted by DPR. Essentially, as illustrated in Figure \ref{fig:sample}, each passage $p$ is composed of multiple sentences, upon which multiple semantically faraway questions can be derived, which forms a question set $\mathcal{Q} = \{q_1, q_2, ..., q_k\}$. Under our investigation, such a \textit{one-to-many problem} is causing severe conflicting problems in the current contrastive learning framework, which we refer to as \textit{Contrastive Conflicts}. To the best of our knowledge, this is the first work that formally studies the conflicting problems in the contrastive learning framework of dense passage retrieval. Here, we distinguish two kinds of  \textit{Contrastive Conflicts}.

\begin{figure}
	\centering
	\includegraphics[width=1.0\linewidth]{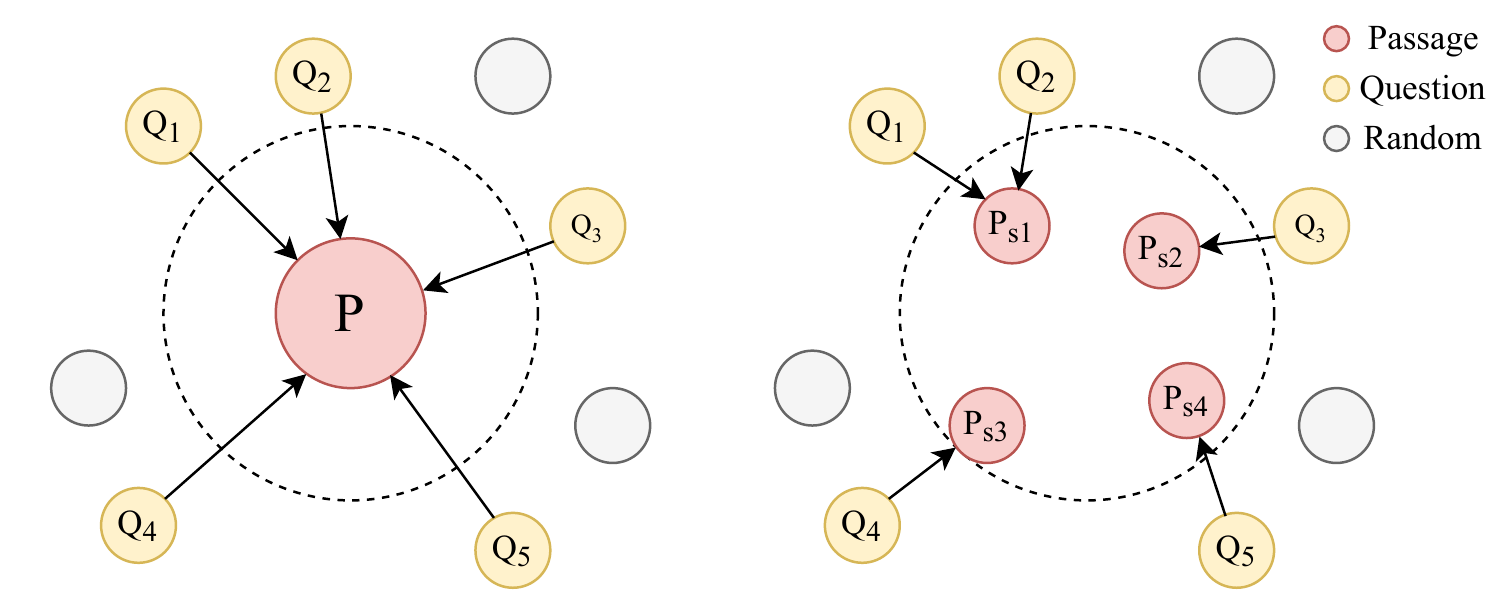}
	\caption{Visualization of contrastive conflicts in DPR (left) and solution provided by our method (right).}
	\label{fig:conf_vis}
\end{figure}

\noindent$\bullet$ \textbf{Transitivity of Similarity} The goal of the contrastive learning framework in DPR is to maximize the similarity between the representation of the question and its corresponding gold passage. As illustrated in Figure \ref{fig:conf_vis}, under \textit{Contrastive Conflicts}, the current contrastive learning framework will unintendedly maximize the similarity between different question representations derived from the same passage, even if they might be semantically different, which is exactly the cause of low performance on SQuAD \cite{rajpurkar2016squad} for DPR (SQuAD has an average of 2.66 questions per passage).


\noindent$\bullet$ \textbf{Multiple References in Large Batch Size} According to \citet{karpukhin2020dense}, the performance of DPR highly benefits from large batch size in the contrastive learning framework. However, under \textit{Contrastive Conflicts}, one passage could be the positive passage $p^{+}$ of multiple questions (i.e. the question set $\mathcal{Q}$). Therefore, a large batch size will increase the probability that some questions of $\mathcal{Q}$ might occur in the same batch. With the widely adopted in-batch negative technique \cite{karpukhin2020dense, lee2020learning}, such $p^{+}$ will be simultaneously referred to as both the positive sample and the negative sample for every $q$ in $\mathcal{Q}$, which is logically unreasonable.

Since \textit{one-to-many problem} is the direct cause of both conflicts, this paper presents a simple but effective strategy that breaks down dense passage representations into contextual sentence level ones, which we refer to as \textbf{D}ense \textbf{C}ontextual \textbf{S}entence \textbf{R}epresentation (DCSR). Unlike long passages, it is hard to derive semantically faraway questions from one short sentence. Therefore, by modeling ODPR in smaller units like contextual sentences, we fundamentally alleviate \textit{Contrastive Conflicts} by solving the \textit{one-to-many problem}. Note that we do not simply encode each sentence separately. Instead, we encode the passage as a whole and use sentence indicator tokens to acquire the sentence representations within the passage, to preserve the contextual information. We further introduce the in-passage negative sampling strategy, which samples neighboring sentences of the positive one in the same passage to create hard negative samples. Finally, concrete experiments have verified the effectiveness of our proposed method from both retrieval accuracy and transferability, especially on datasets where \textit{Contrastive Conflicts} are severe.

\noindent\textbf{Contributions} (i) We investigate the defects of the current contrastive learning framework in training dense passage representation in Open-Domain Passage Retrieval. (ii) To handle \textit{Contrastive Conflicts}, we propose to index the Wikipedia corpus using contextual sentences instead of passages. We also propose the in-passage negative sampling strategy in training the contextual sentence representations. (iii) Experiments show that our proposed method significantly outperforms original baseline, especially on datasets where \textit{Contrastive Conflicts} are severe. Extensive experiments also present better transferability of our DCSR, indicating that our method captures the universality of the concerned task datasets.

\section{Related Work}

\noindent\textbf{Open-Domain Passage Retrieval} Open-Domain Passage Retrieval has been a hot research topic in recent years. It requires a system to extract evidence passages for a specific question from a large passage corpus like Wikipedia, and is challenging as it requires both high retrieval accuracy and specifically low latency for practical usage. Traditional approaches like TF-IDF \cite{ramos2003using}, BM25 \cite{robertson2009probabilistic} retrieve the evidence passages based on the lexical match between questions and passages. Although these lexical approaches meet the requirement of low latency, they fail to capture non-lexical semantic similarity, thus performing unsatisfying on retrieval accuracy. 

With recent advances of pretrained language models (PrLMs) like BERT \cite{devlin2019bert}, RoBERTa \cite{liu2019roberta}, a series of neural approaches based on cross-encoders are proposed \cite{vig2019comparison, wolf2019transfertransfo}. Although enjoying satisfying retrieval accuracy, the retrieval latency is often hard to tolerate in practical use. More recently, the Bi-Encoder structure has captured the researchers' attention. With Bi-Encoder, the representations of the corpus at scale can be precomputed, enabling it to meet the requirement of low latency in passage retrieval. \citet{lee2019latent} first proposes to pretrain the Bi-Encoder with \textit{Inverse Cloze Task} (ICT). Later, DPR \cite{karpukhin2020dense} introduces a contrastive learning framework to train dense passage representation, and has achieved impressive performance on both retrieval accuracy and latency. Based on DPR, many works make further improvements either by introducing better sampling strategy \cite{xiong2020approximate,lu2020neural,tang2021improving,qu2021rocketqa} or extra pretraining \cite{sachan2021end}, or even distilling knowledge from cross-encoders \cite{izacard2021distilling, yang2021neural}. 

Our method follows the contrastive learning research line of ODPR. Different from previous works that focus on either improving the quality of negative sampling or using extra pretraining, we make improvements by directly optimizing the modeling granularity with an elaborately designed contrastive learning training strategy.

\noindent\textbf{Contrastive Learning} Contrastive learning recently is attracting researchers' attention in all area. After witnessing its superiority in Computer Vision tasks \cite{chen2020simple, he2020momentum}, researchers in NLP are also applying this technique \cite{wu2020clear, karpukhin2020dense, yan2021consert, giorgi2020declutr, gao2021simcse}. 
For the concern of ODPR, the research lines of contrastive learning can be divided into two types: (i) Improving the sampling strategies for positive samples and hard negative samples. According to \cite{wu2017sampling}, the quality of positive samples and negative samples are of vital importance in the contrastive learning framework. Therefore, many researchers seek better sampling strategies to improve the retrieval performance \cite{xiong2020approximate}. (ii) Improving the contrastive learning framework. DensePhrase \cite{lee2020learning} uses memory bank like MOCO \cite{he2020momentum} to increase the number of in-batch negative samples without increasing the GPU memory usage, and models retrieval process on the phrase level but not passage level, achieving impressive performance. 

Our proposed method follows the second research line. We investigate a special phenomenon, \textit{Contrastive Conflicts} in the contrastive learning framework, and experimentally verify the effectiveness of mediating such conflicts by modeling ODPR in a smaller granularity. More similar to our work, \citet{akkalyoncu-yilmaz-etal-2019-cross} also proposes to improve dense passage retrieval based on sentence-level evidences, but their work is not in the research line of contrastive learning, and focuses more on passage re-ranking after retrieval but not retrieval itself.

\section{Methods}

\begin{figure*}
	\centering
	\includegraphics[width=1.0\linewidth]{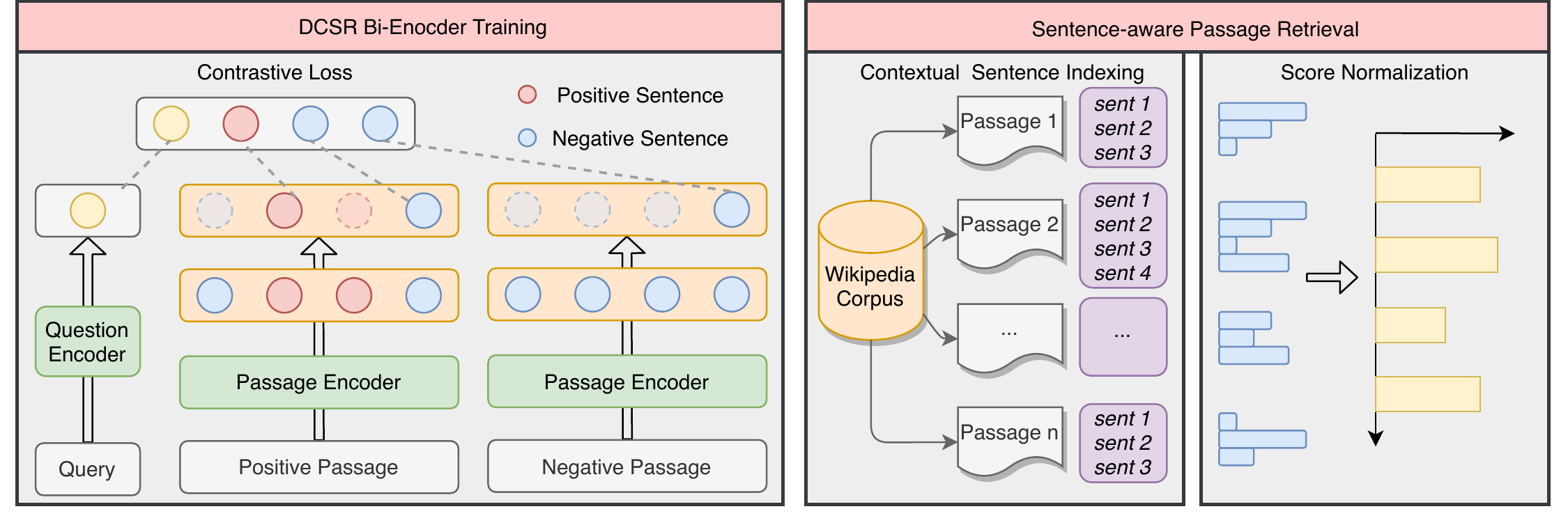}
	\caption{An illustration of our DCSR processing pipeline. The left part shows the contrastive training paradigm of our method, and the right part presents the inference pipeline.}
	\label{fig:DCSR_encoder}
\end{figure*}

\subsection{Contrastive Learning Framework}


Existing contrastive learning framework aims to maximize the similarity between the representations of each question and its corresponding gold passages.

Suppose there is a batch of $n$ questions, $n$ corresponding gold passages and in total $k$ hard negative passages. Denote the questions in batch as $q_1, q_2, ..., q_n$, their corresponding gold passages as $gp_1, gp_2, ..., gp_n$, and hard negative passages as $np_1, np_2, ..., np_k$. Two separate PrLMs are first used separately to acquire representations for questions and passages $\{h_{q_1}, h_{q_2}, ...; h_{gp_1}, h_{gp_2}, ...; h_{np_1}, h_{np_2}, ...\}$. The training objective for each question sample $q_{i}$ of original DPR is shown in Eq (\ref{eq_dpr}):

\begin{equation} \label{eq_dpr}
\begin{aligned}
& \mathcal{L}\left(q_{i}, gp_{1}, \cdots, gp_{n}, np_{1}, \cdots, np_{k}\right) = \\
&-\log \frac {e^{\operatorname{sim}\left(h_{q_{i}}, h_{gp_{i}}\right)}} {{\sum_{j=1}^{n} e^{\operatorname{sim}\left(h_{q_{i}},h_{gp_{j}}\right)}} + \sum_{j=1}^{k} e^{\operatorname{sim}\left(h_{q_{i}}, h_{np_{j}}\right)}}
\end{aligned}
\end{equation}
The $\operatorname{sim}(\cdot)$ could be any similarity operator that calculates the similarity between the question representation $h_{q_{i}}$ and the passage representation $h_{p_{j}}$. 

Minimizing the objective in Eq (\ref{eq_dpr}) is the same as (i) maximizing the similarity between each  $h_{q_{i}}$ and $h_{gp_{i}}$ pair, and (ii) minimizing the similarity between  $h_{q_{i}}$ and all other $h_{gp_{j}}$ ($i \neq j$) and $h_{np_{k}}$. As discussed previously, this training paradigm will cause conflicts under current contrastive learning framework due to (i) Transitivity of Similarity, and (ii) Multiple References in Large Batch Size.


\subsection{Dense Contextual Sentence Representation}

The cause of the \textit{Contrastive Conflicts} lies in \textit{one-to-many problem}, that most of the passages are often organized by multiple sentences, while these sentences may not always stick to the same topic, as depicted in Figure \ref{fig:sample}. Therefore, we propose to model passage retrieval in a smaller granularity, i.e. contextual sentences, to alleviate the occurrence of \textit{one-to-many problem}. 

Since contextual information is also important in passage retrieval, simply breaking down passages into sentences and encoding them independently is infeasible. Instead, following \cite{beltagy2020longformer, lee2020slm, wu2021graph}, we insert a special $<$sent$>$ token at the sentence boundaries in each passage, and encode the passage as a whole to preserve the contextual information, which results in the following format of input for each passage:
$$
\centerline{\textup{[CLS] $<$sent$>$ sent$_1$ $<$sent$>$  sent$_2$ ... [SEP]}}
$$

We then use BERT \cite{devlin2019bert} as encoder to get the contextual sentence representations by these indicator $<$sent$>$ tokens. For convenience of illustration, taking a give query $q$ into consideration, we denote the corresponding positive passage in the training batch as $p^{+}$, which consists of several sentences:
$$\mathcal{P} = \{p_{s_1^{-}}, p_{s_2^{-}}, ... p_{s_i^{+}}, ... p_{s_{k-1}^{-}}, p_{s_{k}^{-}}\}$$

Similarly, we denote the corresponding BM25 negative passage as: 
$$\mathcal{N} = \{n_{s_1^{-}}, n_{s_2^{-}}, ... n_{s_i^{-}}, ... n_{s_{k-1}^{-}}, n_{s_{k}^{-}}\}$$
Here $(*)^{-/+}$ means whether the sentence or passage contains the gold answer. We refine the original contrastive learning framework by creating sentence-aware positive and negative samples. The whole training pipeline is shown in the left part of Figure \ref{fig:DCSR_encoder}.

\subsubsection{Positives and Easy Negatives}  Following \citet{karpukhin2020dense}, we use BM25 to retrieve hard negative passages for each question. To build a contrastive learning framework based on contextual sentences, we consider the sentence that contains the gold answer as the positive sentence (i.e. $p_{s_i^{+}}$), and randomly sample several negative sentences (random sentences from $\mathcal{N}$) from a BM25 random negative passage. Also, following \cite{karpukhin2020dense, lee2020learning}, we introduce in-batch negatives as additional easy negatives.

\subsubsection{In-Passage Negatives} To handle the circumstance where multiple semantically faraway questions may be derived from one single passage, we hope to encourage the passage encoder to generate contextual sentence representations as diverse as possible for sentences in the same passage. Noticing that not all the sentences in the passage contain the gold answer and stick to the topic related to the given query, we further introduce in-passage negatives to maximize the difference between contextual sentences representations within the same passage. Concretely, we randomly sample one sentence that does not contain the gold answer (i.e. a random sentence from $\mathcal{P}/\{P_{s_i^{+}}\}$). Note that a positive passage might not contain such sentence. If it does not exist, this in-passage negative sentence is substituted by another easy negative sentence from the corresponding BM25 negative passage (a random sentence from $\mathcal{N}$). These in-passage negatives function as hard negative samples in our contrastive learning framework.


\subsection{Retrieval}

For retrieval, we first use FAISS \cite{johnson2019billion} to calculate the matching scores between the question and all the contextual sentence indexes. As one passage has multiple keys in the indexes, we retrieve top $100 \times k$ ($k$ is the average number of sentences per passage) contextual sentences for inference. To change these sentence-level scores into passage-level ones, we adopt a probabilistic design for ranking passages, which we refer to as Score Normalization. 

\noindent\textbf{Score Normalization} After getting the scores for each contextual sentences to each question by FAISS, we first use a Softmax operation to normalize all these similarity scores into probabilities. Suppose one passage $\mathcal{P}$ with several sentences $s_1, s_2, ..., s_n$, and denote the probability for each sentence that contains the answer as $p_{s_1}, p_{s_2}, ..., p_{s_n}$, we can calculate the probability that \textit{the answer is in passage $\mathcal{P}$} by Equation \ref{eq:hasans}.
\begin{equation} \label{eq:hasans}
HasAns(\mathcal{P}) = 1 - \prod \limits_{i=1}^n (1-p_{s_i}) 
\end{equation}

We then re-rank all the retrieved passages by $HasAns(\mathcal{P})$, and select the top 100 passages for evaluation in our following experiments.


\section{Experiments}



\subsection{Datasets}

\noindent\textbf{OpenQA Dataset} OpenQA \cite{lee2019latent} collects over 21 million 100-token passages from Wikipedia to simulate the open-domain passage corpus. OpenQA also collects question-answer pairs from existing datasets, including SQuAD \cite{rajpurkar2016squad}, TriviaQA \cite{joshi2017triviaqa}, Natural Questions \cite{kwiatkowski2019natural}, WebQuestions \cite{berant2013semantic} and TREC \cite{baudivs2015modeling}.

\begin{table}[tp]
	\centering
	\setlength{\tabcolsep}{6pt}
	\fontsize{10.0pt}{\baselineskip}\selectfont
	\begin{tabular}{lccccc}
		\toprule
		& 1 & 2 & 3 & $\geq4$ & Avg\\
		\midrule
		SQuAD & \ \;8,482 & 6,065 & 5,013 & 6,754 & 2.66  \\
		Trivia & 43,401 & 5,308 & 1,206 & \ \;587 & 1.20 \\
		NQ & 32,158 & 4,971 & 1,670 & 1,871 & 1.45\\
		\bottomrule
	\end{tabular}
	\caption{Occurrence of \textit{one-to-many problem} in training sets.}
	\label{tab:one-to-many}
\end{table}

\begin{table*}[tp]
	\centering
	\setlength{\tabcolsep}{4.0pt}
	\fontsize{10.0pt}{\baselineskip}\selectfont
	\begin{tabular}{lL{1.65cm}L{1.65cm}L{1.65cm}L{1.65cm}L{1.65cm}L{1.65cm}}
		\toprule
		\multirow{2}{*}{Model} & \multicolumn{3}{c}{Top-20} & \multicolumn{3}{c}{Top-100}\\
		& NQ & Trivia & SQuAD & NQ & Trivia & SQuAD \\
		\midrule
		\multicolumn{7}{l}{\textit{Base Architecture Comparison -- Single}} \\
		DPR \cite{karpukhin2020dense}  & 78.4 & 79.4 & 52.8$\dagger$ & 85.4 & 85.0 & 71.0$\dagger$ \\
		
		DCSR (Ours) & \textbf{78.9}(+0.5) & \textbf{79.7}(+0.3) & \textbf{63.7}(+10.9) & \textbf{86.5}(+1.1) & \textbf{85.2}(+0.2) & \textbf{78.1}(+7.1)  \\
		
		\hdashline
		\multicolumn{7}{l}{\textit{Base Architecture Comparison -- Multi}} \\
		DPR \cite{karpukhin2020dense}  & \textbf{79.4} & 78.8 & 51.6 & 86.0 & 84.7 & 67.6 \\
		DCSR (Ours) & 79.1(-0.3) & \textbf{79.6}(+0.8) & \textbf{63.8}(+12.2) & \textbf{86.6}(+0.6) & \textbf{85.2}(+0.5) & \textbf{77.6}(+10.0)  \\
		
		\bottomrule
	\end{tabular}
	\caption{Retriever Performance Comparison on the test sets. ``$\dagger$": For SQuAD dataset on DPR in the \textit{Single} setting, we are not able to reproduce the original results from the official DPR code\footnotemark[1]\footnotemark[2]. Instead, we rerun DPR on SQuAD in the \textit{Single} setting and report its performance based on our reproduction. The parameter settings are shared between our DPR reproduction and DCSR to ensure fairness. Other statistics are taken from \citet{karpukhin2020dense}.}
	\label{tab:retriev_cmp}
\end{table*}

We experiment our proposed method on SQuAD, TriviaQA and NQ. For the previously concerned \textit{Contrastive Conflicts} problem, we also analyze the existence frequency of the conflicting phenomenon for each dataset. We count the number of questions for each passage, i.e, the times that this passage is referred to as the positive sample. The corresponding results are shown in Table \ref{tab:one-to-many}. From this table, we can see that of all three datasets we choose, SQuAD is most severely affected by the \textit{Contrastive Conflicts} problem, that many passages occur multiple times as the positive passages for different questions. These statistics are consistent with the fact that DPR performs the worst on SQuAD, while acceptable on Trivia and NQ.

\subsection{Training and Implementation Details} 

\noindent\textbf{Hyperparameters} In our main experiments, we follow the hyperparameter setting in DPR \cite{karpukhin2020dense} to acquire comparable performance, i.e. an initial learning rate of 2e-5 for 40 epochs on each dataset. We use 8 Tesla V100 GPUs to train the Bi-Encoder with a batch size of 16 on each GPU.

\noindent\textbf{Extra Cost} Although we are modeling passage retrieval in a totally different granularity, our method adds little extra computation overhead compared to DPR. For model complexity, our proposed method adopts exactly the same model structure as DPR does, meaning that there are no additional parameters introduced. For training time, the negative sentences in our method are randomly sampled from the negative passage in DPR. Therefore, the extra time burden brought by our method is only caused by the sampling procedure, which is negligible. 

\noindent\textbf{Training Settings} To have a comprehensive comparison with DPR, we train DCSR under three different settings. (i) \textit{Single}, where each dataset is both trained and evaluated under their own domain. (ii) \textit{Multi}, where we use a combination of the NQ, Trivia and SQuAD datasets to train a universal Bi-Encoder, and evaluate its performance on the test sets of all three datasets. (iii) \textit{Adversarial Training}, which is a simple negative sampling strategy. We first use the original dataset to train a DPR or DCSR checkpoint, and use such checkpoint to acquire semantically hard negative passages from the whole Wikipedia corpus.

\footnotetext[1]{Code in \url{https://github.com/facebookresearch/DPR}.}

\footnotetext[2]{It is an issue that is shared by researchers on \href{https://github.com/facebookresearch/DPR/issues/123}{github}. More discussion about this result will be discussed in Appendix \ref{append:squad}.}

\subsection{Main Results on Passage Retrieval}
Table \ref{tab:retriev_cmp} shows our main results on OpenQA. 

\noindent\textbf{For the \textit{Single} setting}, (i) Consistent with the core aim of this paper that our proposed sentence-aware contrastive learning solves \textit{Contrastive Conflicts}, DCSR achieves significantly better results than DPR especially on the dataset that is severely affected by \textit{Contrastive Conflicts}. For example, on the SQuAD dataset, our method achieves 10.9\% performance gain on the Top-20 metric, and 7.1\% performance gain on the Top-100 metric. (ii) For datasets that are less affected by \textit{Contrastive Conflicts}, like NQ and Trivia, we still achieve slight performance gain on all metrics.

\noindent\textbf{For the \textit{Multi} setting}, DPR on Trivia and SQuAD suffers from a significant performance drop compared to \textit{Single} setting, while our model is only slightly affected. It indicates that our proposed sentence-aware contrastive learning not only solves the \textit{Contrastive Conflicts}, but also captures the universality of datasets from different domains.

\begin{table}[tp]
    \centering
	\setlength{\tabcolsep}{5.0pt}
	\fontsize{10.0pt}{\baselineskip}\selectfont
    \begin{tabular}{llcccc}
        \toprule
		\multirow{2}{*}{Model} & & \multicolumn{2}{c}{Top-20} & \multicolumn{2}{c}{Top-100}\\
		& & NQ & Trivia & NQ & Trivia \\
		\midrule
        DPR & + adv-train & 81.3 & - & 87.3 & -  \\
		& + ANCE
		& \textbf{81.9} & \textbf{80.3} & \textbf{87.5} & 85.3 \\
		& \multicolumn{2}{c}{\cite{xiong2020approximate}} & & & \\
		
		
		
		\hdashline
		DCSR & + adv-train & 81.4 & 80.0 & \textbf{87.5} & \textbf{85.7} \\
	\bottomrule
    \end{tabular}
    \caption{Performance Comparison when incorporated with negative sampling strategy.}
    \label{tab:negative_sample}
\end{table}


\begin{table*}[tp]
	\centering
	\setlength{\tabcolsep}{5.0pt}
	\fontsize{10.0pt}{\baselineskip}\selectfont
	\begin{tabular}{llC{1.2cm}C{1.2cm}C{1.2cm}C{1.2cm}C{1.2cm}C{1.2cm}}
		\toprule
		\multirow{2}{*}{Model} & & \multicolumn{3}{c}{Top-20} & \multicolumn{3}{c}{Top-100} \\
		& & NQ & Trivia & SQuAD & NQ & Trivia & SQuAD \\
		\midrule
		DPR  & \cite{karpukhin2020dense} & 43.7  & 62.1 & 46.5  & 54.0 & 72.4 & 63.6 \\
		
		DCSR & + 1 BM25 random   & 44.5 & 63.1 & 51.1 & 54.5 & 72.9 & 66.6 \\
		
		& + 2 BM25 random & 44.0 & \textbf{63.5} & 50.3 & 54.7 & 72.9 & 65.1 \\
		
		& + 1 in-passage \& +1 BM25 random & \textbf{45.2} & 63.4 & \textbf{54.5} & \textbf{55.3} & \textbf{73.2} & \textbf{68.5} \\
		
		\bottomrule
	\end{tabular}
	\caption{Ablations of Negative Sampling Strategy on Wikipedia subset (1/20 of the whole corpus) in the \textit{Single} Setting.}
	\label{tab:ablation}
\end{table*}


\subsection{Incorporated with Negative Sampling}

Different from other frontier researches which mainly devote themselves either to investigating better negative sampling strategies, like ANCE \cite{xiong2020approximate}, NPRINC \cite{lu2020neural}, etc., or to extra pretraining \cite{sachan2021end}, or to distilling knowledge from cross-encoders \cite{izacard2021distilling, yang2021neural}, our proposed method directly optimizes the modeling granularity in DPR. Therefore, our method could be naturally incorporated with these researches and achieve better results further. Due to computational resource limitation, we do not intend to replicate all these methods, but use \textit{adversarial training} as an example. Following ANCE \cite{xiong2020approximate}, we conduct experiments on NQ and Trivia to show the compatibility of our method, listed in Table \ref{tab:negative_sample}. With such a simple negative sampling strategy, our DCSR achieves comparable results with its DPR counterpart.

\subsection{Ablation Study}

To illustrate the efficacy of the previously proposed negative sampling strategy, we conduct an ablation study on a subset of OpenQA Wikipedia corpus\footnote{Because evaluating on the whole Wikipedia corpus takes too much resource and time (over 1 day per experiment per dataset).}. We sample 1/20 of the whole corpus, which results in a collection of 1.05 million passages in total. As reference, we reproduce DPR and also list their results in Table \ref{tab:ablation}. We compare the following negative sampling strategies of our proposed method.

\noindent\textbf{+ 1 BM25 random} In this setting, we randomly sample (i) one gold sentence from the positive passage as the positive sample, and (ii) one negative sentence from the negative passage as the negative sample per question.

\noindent\textbf{+ 2 BM25 random} In this setting, we randomly sample (i) one gold sentence from the positive passage as the positive sample, and (ii) two negative sentences from two different negative passages as two negative samples per question.

\noindent\textbf{+ 1 in-passage \& + 1 BM25 random} In this setting, we randomly sample (i) one gold sentence from the positive passage as the positive sample, (ii) one negative sentence from the positive passage as the first negative sample, and (iii) one negative sentence from the negative passage as the second negative sample per question.

\noindent\textbf{Ablations of Negative Sampling Strategy} The results are shown in Table \ref{tab:ablation}. (i) Under the circumstance where only 1.05 million passages are indexed, variants of our DCSR generally perform significantly better than DPR baseline, especially on NQ dataset (over 1\% improvement on both Top-20 and Top-100) and SQuAD dataset (8.0\% improvement on Top-20 and 4.9\% improvement on Top-100), which verifies the effectiveness of solving \textit{Contrastive Conflicts}. (ii) Further, we found that increasing the number of negative samples helps little, but even introduces slight performance degradation on several metrics. (iii) The in-passage negative sampling strategy consistently helps in boosting the performance of nearly all datasets on all metrics, especially on the SQuAD dataset, which is consistent with our motivation for in-passage negatives, which is to encourage a diverse generation of contextual sentence representations within the same passage in solving the \textit{one-to-many problem}.

\noindent\textbf{Ablations of Training Data} The results are shown in Table \ref{tab:abl_negsample}. (i) We first directly use the augmented adversarial training dataset provided by DPR (marked as \textit{DPR-hard}) and train our DCSR, having achieved even better results on the NQ dataset. This augmented dataset is sub-optimal for our model, as these hard negative samples are passage-specific, while our model prefers sentence-specific ones. (ii) We then use our previous best DCSR checkpoint to retrieve a set of sentence-specific hard negatives (marked as \textit{DCSR-hard}) and train a new DCSR, which achieves further performance gain on both metrics on NQ dataset.

\begin{table}[tp]
    \centering
	\setlength{\tabcolsep}{7.2pt}
	\fontsize{10.0pt}{\baselineskip}\selectfont
    \begin{tabular}{lcccc}
        \toprule
		\multirow{2}{*}{Model} & \multicolumn{2}{c}{Top-20} & \multicolumn{2}{c}{Top-100}\\
		& NQ & Trivia & NQ & Trivia \\
		\midrule
		DPR$_{raw-data}$ & 43.7 & 62.1 & 54.0 & 72.4 \\
		DPR$_{\textit{DPR-hard}}$ & 47.6 & - & 56.5 & -  \\
		DCSR$_{\textit{DPR-hard}}$ & 47.6 & -  & 57.0 & -  \\
		DCSR$_{\textit{DCSR-hard}}$ & \textbf{48.8} & \textbf{66.2}  & \textbf{57.1} &\textbf{75.0} \\
	\bottomrule
    \end{tabular}
    \caption{Ablations of Training Data. For Trivia, \textit{DPR-hard} is not provided in the original paper.}
    \label{tab:abl_negsample}
\end{table}

\begin{table*}[tp]
	\centering
	\setlength{\tabcolsep}{6.5pt}
	\fontsize{10.0pt}{\baselineskip}\selectfont
	\begin{tabular}{lC{1.8cm}C{0.9cm}C{1.8cm}C{0.9cm}C{1.8cm}C{0.9cm}C{1.8cm}C{0.9cm}}
		\toprule
		& \multicolumn{4}{c}{SQuAD-to-Trivia} & \multicolumn{4}{c}{NQ-to-Trivia} \\
		Model & Top 20 & diff & Top 100 & diff & Top 20 & diff & Top 100 & diff \\
		\midrule
		DPR & 48.7/62.1 & $\downarrow$13.4 & 64.5/72.4 & $\downarrow$7.9 & 48.8/62.1 & $\downarrow$13.3 & 62.7/72.4 & $\downarrow$9.7 \\
		DCSR & 54.0/63.4 & $\downarrow$\textbf{9.4} & 67.8/73.2 & $\downarrow$\textbf{5.4} & 52.7/63.4 & $\downarrow$\textbf{10.7} & 65.9/73.2 & $\downarrow$\textbf{7.3} \\
		\bottomrule
	\end{tabular}
	\caption{Transferability comparing our methods with DPR. We train the retriever model on the SQuAD dataset or the NQ dataset, and evaluate it on Trivia QA (statistics on the left). For reference, we also list the performance where the retriever model is both trained and evaluated on the Trivia QA (statistics on the right).}
	\label{tab:transfer_cmp_v2}
\end{table*}

\section{Discussion}
In this section, we discuss the transferability difference and the influence of Wikipedia corpus size on both DPR and our DCSR. More discussions from different aspects are presented in the Appendices, including (i) Validation accuracy on dev sets in Appendix \ref{sec:valid_acc}, which is also a strong evidence of alleviating \textit{Contrastive Conflicts}. (ii) Error analysis for SQuAD in Appendix \ref{append:squad}, which further shows the generalization ability of our method. (iii) Case study in Appendix \ref{case_study}, which discusses the future improvement of DCSR.

\subsection{Transferability}

To further verify that our learned DCSR is more suitable in Open-Domain Passage Retrieval, especially under the \textit{Contrastive Conflicts} circumstance, we conduct experiments to test the transferability between DPR and our DCSR. Similarly, instead of running such experiments on the entire Wikipedia corpus, we sample 1/20 of the corpus, which results in a collection of 1.05 million passages in total. We test the transferability result from SQuAD to Trivia and from NQ to Trivia, as compared to Trivia, both SQuAD and NQ suffer more from \textit{Contrastive Conflicts}. The results are shown in Table \ref{tab:transfer_cmp_v2}.

From Table \ref{tab:transfer_cmp_v2}, when compared to DPR, our model enjoys significantly better transferability. In both scenarios, DPR shows over 2\% performance gap in all metrics of the transferability tests, indicating that our method performs much better in generalization across the datasets. This phenomenon once again confirms our theorem, that by modeling passage retrieval in the granularity of contextual sentences, our DCSR well models the universality across the datasets, and shows much better transferability than DPR.
	
\subsection{Corpus Size}\label{corpus_size}

In our extensive experiments, we further found out that our method can achieve overwhelming better performance than DPR on smaller corpus. In this experiment, we take \textit{the first 0.1 million}, \textit{the first 1.05 million} and \textit{all passages} from the original Wikipedia corpus, and conduct 
dense retrieval on these three corpora varied in size. The statistic results are shown in Table \ref{tab:wiki_size}. 


From Table \ref{tab:wiki_size}, first of all, our model achieves better performance than DPR in all settings, where such improvement is more significant in smaller corpus. On the setting where only 0.1 million passages are indexed in the corpus, our model achieves over 2.0\% exact improvement on all metrics on both NQ and Trivia. We speculate this is because of the following two strengths of our method. 

\noindent$\bullet$ The alleviation of \textit{Contrastive Conflicts}, which we have analyzed previously. 

\noindent$\bullet$ Modeling passage retrieval using contextual sentences enables a diverse generation of indexes. Some sentences may not be the core aim of their corresponding passages, but can still be the clue for some questions.

Secondly, we can discover that the performance gap between DPR and DCSR is decreasing when the size of Wikipedia corpus increases. This is because with the expansion of indexing corpus, many questions that cannot be solved in the small corpus setting may find much more closely related passages in the large corpus setting, which gradually neutralizes the positive effect brought by the second strength of our proposed method discussed above. Still, our model achieves better performance under the full Wikipedia setting on all datasets and all metrics.

\begin{table}[tp]
	\centering
	\setlength{\tabcolsep}{4.0pt}
	\fontsize{10.0pt}{\baselineskip}\selectfont
	\begin{tabular}{lC{1.0cm}C{1.0cm}C{1.0cm}C{1.0cm}C{1.0cm}C{1.0cm}c}
		\toprule
		\multirow{2}{*}{Model} & \multicolumn{2}{c}{Top-20} & \multicolumn{2}{c}{Top-100} & \multirow{2}{*}{Wiki} \\
		& NQ & Trivia & NQ & Trivia & \\
		\midrule
		
	    DPR  & 25.5 & 39.4 & 36.7 & 51.9 & \multirow{3}{*}{0.10M} \\
		
		DCSR   & \textbf{27.8} & \textbf{41.0} & \textbf{39.0} & \textbf{53.6} &   \\
		
		$\Delta$   & +2.3 & +1.6 & +2.3 & +1.7 &   \\
		
		\hdashline
		
		DPR  & 43.7  & 62.1 & 54.0 & 72.4 & \multirow{3}{*}{1.05M} \\
		
		DCSR   & \textbf{45.2} & \textbf{63.4} & \textbf{55.3} & \textbf{73.2} &   \\
		
		$\Delta$   & +1.5 & +1.3 & +1.3 & +0.8 &   \\
		
		\hdashline
		DPR  & 78.4 & 79.4 & 85.4 & 85.0 & \multirow{3}{*}{21.0M} \\
		
		DCSR   & \textbf{78.9} & \textbf{79.7} & \textbf{86.5} & \textbf{85.2} &   \\
		
		$\Delta$ & +0.5 & +0.3 & +1.1 & +0.2 &  \\
		\bottomrule
	\end{tabular}
	\caption{Retrieval performance when the size of Wikipedia Corpus is varied.}
	\label{tab:wiki_size}
\end{table}

\section{Conclusion}
In this paper, we make a thorough analysis on the \textit{Contrastive Conflicts} issue in the current open-domain passage retrieval. To well address the issue, we propose an enhanced sentence-aware conflict learning method by carefully generating sentence-aware positive and negative samples. We show that the dense contextual sentence representation learned from our proposed method achieves significant performance gain compared to the original baseline, especially on datasets with severe conflicts. Extensive experiments show that our proposed method also enjoys better transferability, and well captures the universality in different datasets.

\clearpage

\bibliographystyle{acl_natbib}

\clearpage

\begin{figure*}[ht]
	\centering
	\subfigure{
		\begin{minipage}[t]{0.32\linewidth}
		    \centering
			\includegraphics[width=0.96\linewidth]{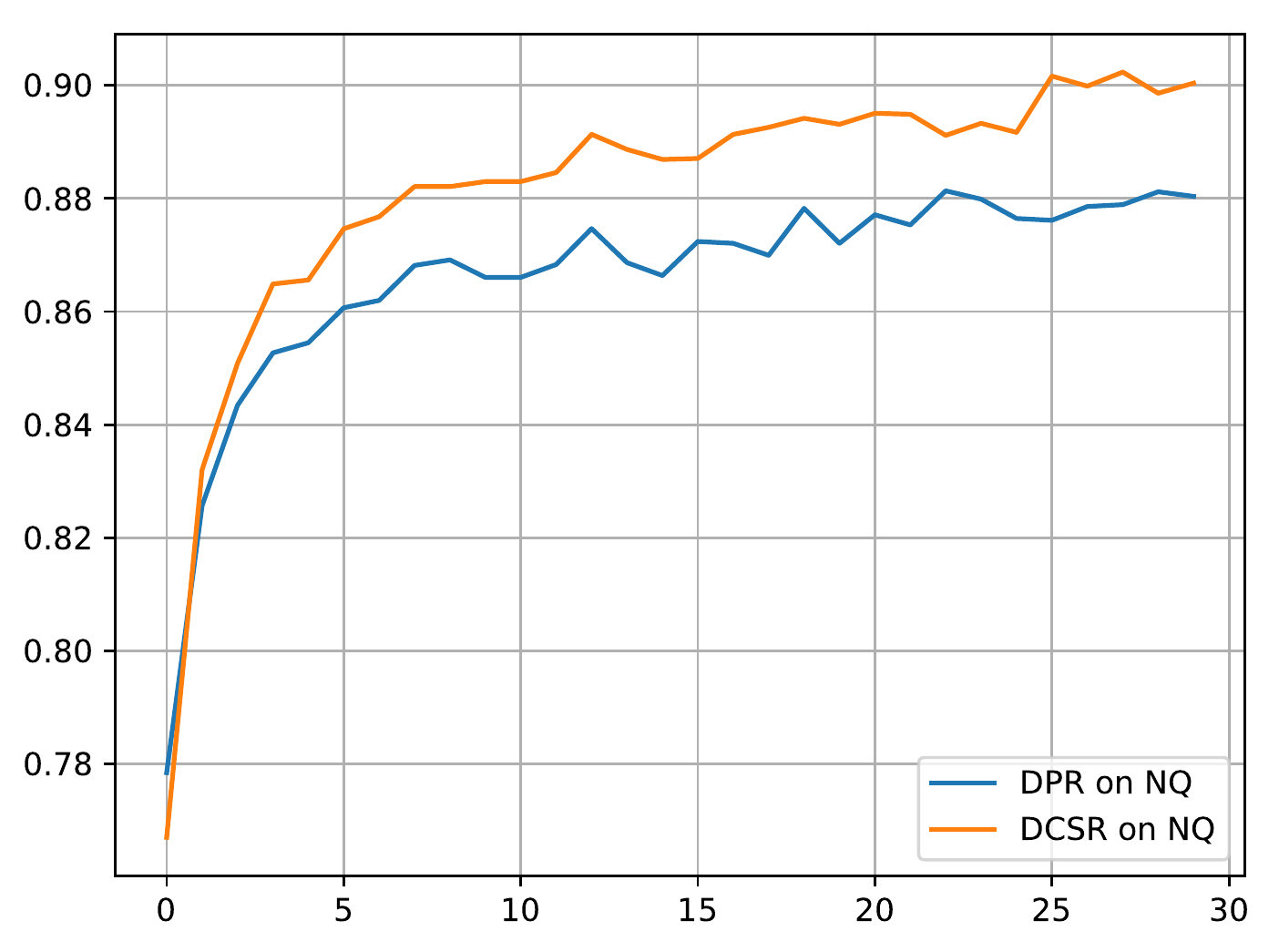}
		\end{minipage}
	}%
	\subfigure{
		\begin{minipage}[t]{0.32\linewidth}
			\centering
			\includegraphics[width=0.96\linewidth]{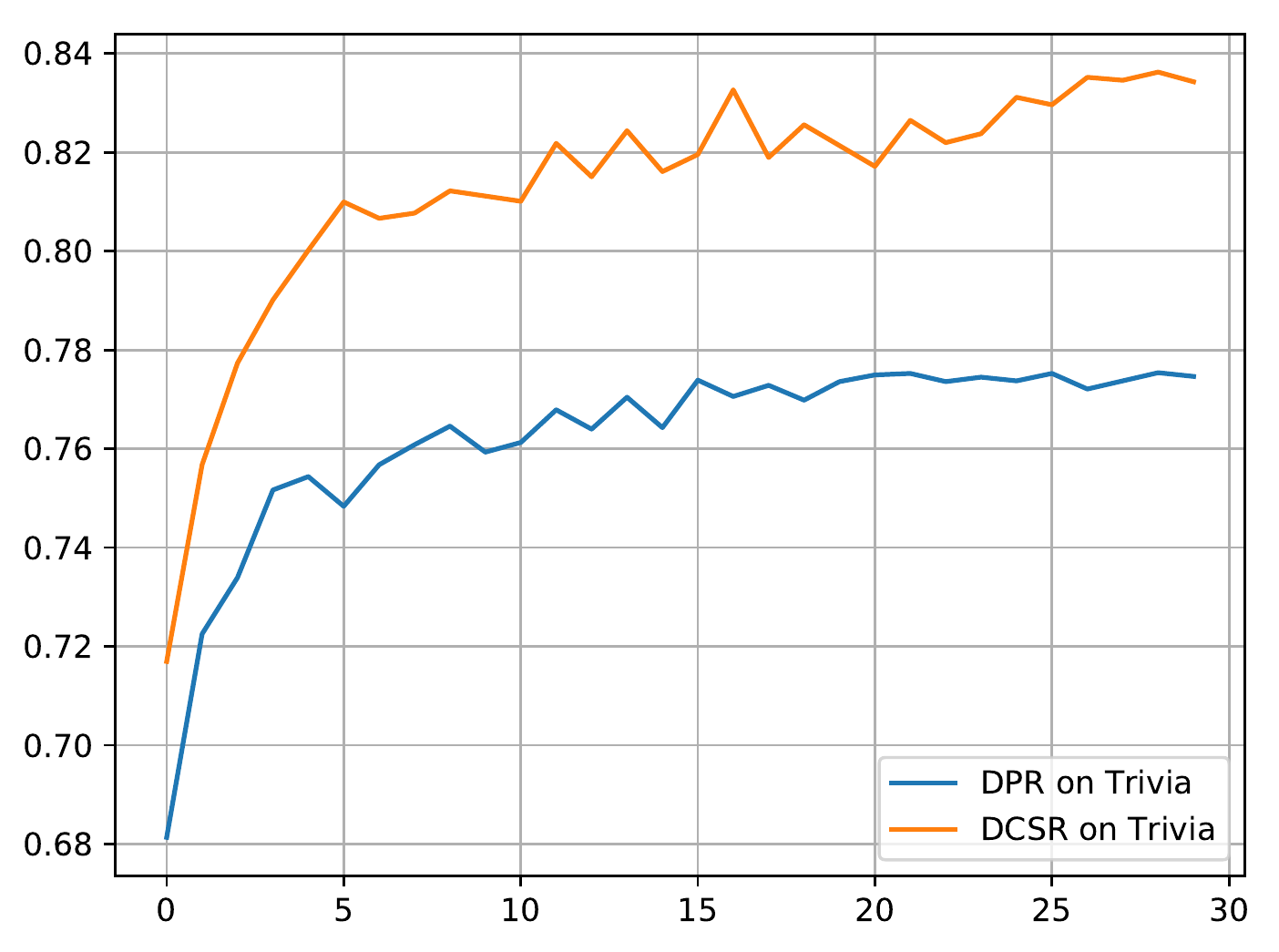}
		\end{minipage}
	}%
	\subfigure{
		\begin{minipage}[t]{0.32\linewidth}
			\centering
			\includegraphics[width=0.96\linewidth]{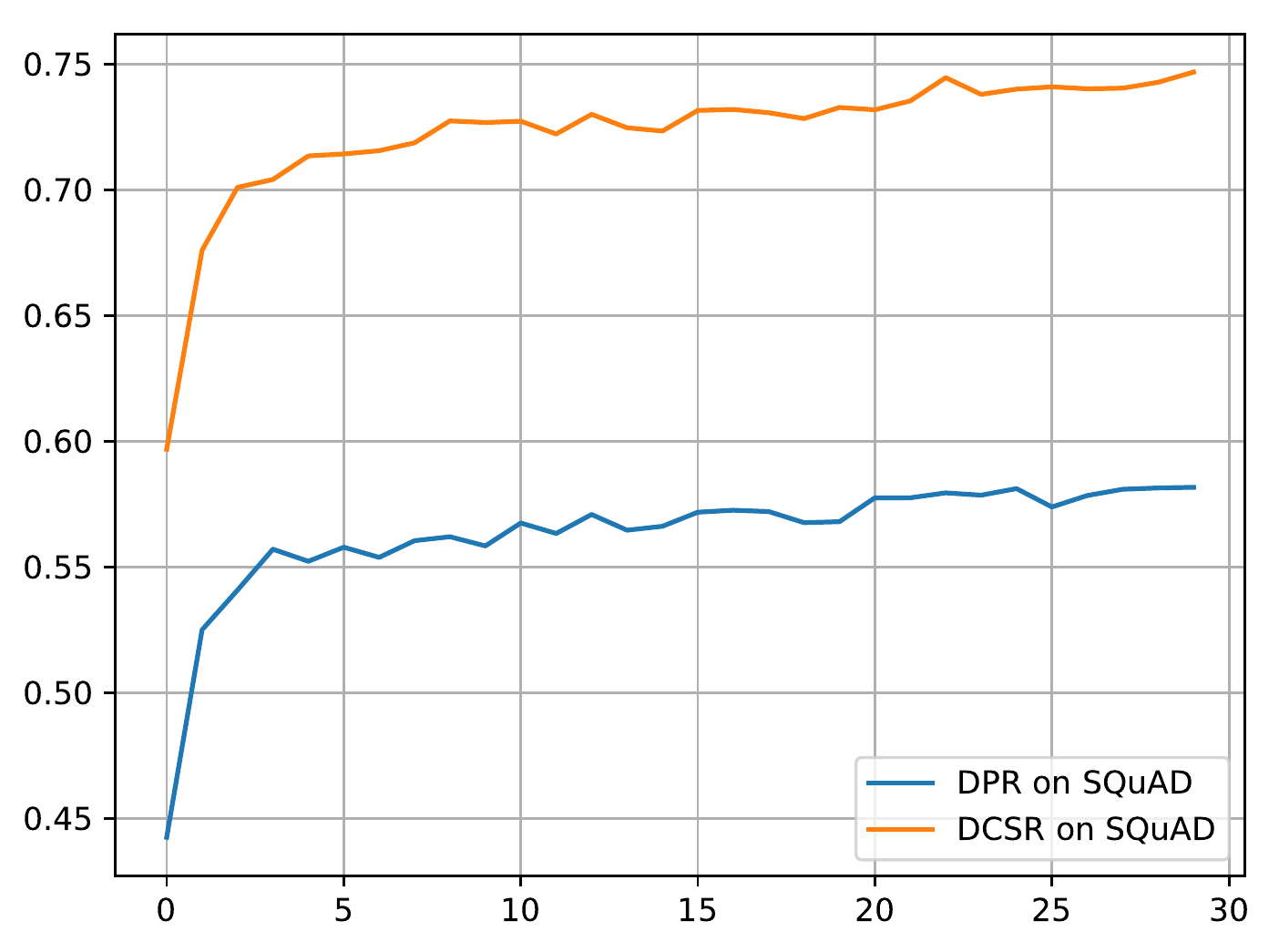}
		\end{minipage}
	}%
	\caption{Dev accuracy in training the encoder of DCSR, including NQ (left), Trivia (middle) and SQuAD (right).}
	\label{fig:dev_acc}
\end{figure*}

\appendix

\begin{figure*}[htb]
	\centering
	\includegraphics[width=0.98\linewidth]{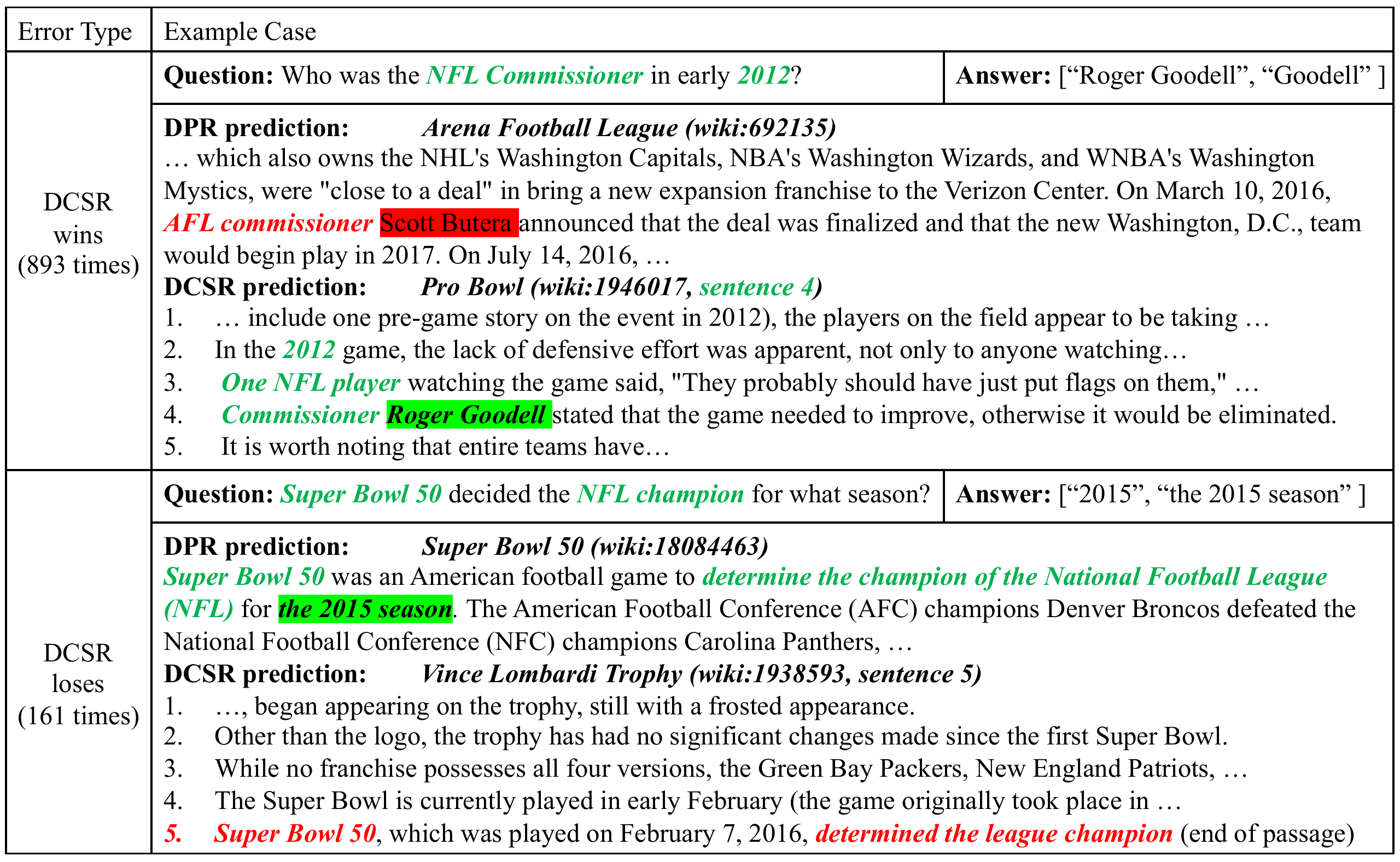}
	\caption{Error Case Study of Our DCSR on SQuAD. Green color represents the correct clues and correct answers, while red color represents wrong ones.}
	\label{tab:case_study}
\end{figure*}

\section{Validation Accuracy} \label{sec:valid_acc}

One may argue that the improvement of DCSR might be due to the expansion of indexing corpus (which we have discussed in previous sections), but not the alleviation of \textit{Contrastive Conflicts}. In this section, we present the validation accuracy comparison during the training process between DPR and our DCSR, which is a strong evidence that DCSR well handles the problem of \textit{Contrastive Conflicts}. 

Under 8 V100 GPUs with a batch size of 16 on each GPU, the validation process could be viewed as a tiny retrieval process for both DPR and DCSR. To maintain a similar validation environment for fair comparison, we use the \textit{+1 BM25 random} version of DCSR, which results in 8*16=128 questions and 2*8*16=256 contextual sentences in one batch. Therefore, the validation process could be interpreted as \textit{retrieving the most relevant contextual sentence for each question in a corpus of 256 sentences}. Under such a validation task, the size of the indexing corpus is restricted to the same for both DPR and DCSR.

The result is shown in Figure \ref{fig:dev_acc}. For both Trivia and NQ, DCSR performs consistently better than DPR with a small accuracy margin. On SQuAD, especially, our DCSR can achieve higher validation accuracy than DPR with only one single epoch, and achieves nearly 20\% final validation accuracy improvement. This phenomenon further verifies that improvement of DCSR is also achieved by improving the training strategy which alleviates \textit{Contrastive Conflicts}, but not only the expansion of the indexing corpus.

\section{Error Analysis for SQuAD} \label{append:squad}


Although achieving overwhelmingly better performance on SQuAD than DPR, our DCSR on SQuAD still lags far behind its counterparts on NQ or Trivia. Interestingly, we found that the results on SQuAD dev sets are pretty good and comparable to the results on NQ or Trivia. The results of both DPR and DCSR on dev set and test set performance are shown in Table \ref{tab:squad_dev_result}.

\begin{table}[tp]
    \centering
	\setlength{\tabcolsep}{8.0pt}
	\fontsize{10.0pt}{\baselineskip}\selectfont
    \begin{tabular}{lcccc}
    \toprule
    \multirow{2}{*}{Model} & \multicolumn{4}{c}{SQuAD-dev} \\
     & Top-1 & Top-5 & Top-20 & Top-100 \\
    \midrule
    DPR & 15.8 & 34.5 & 52.8 & 71.0 \\
    DCSR & \textbf{26.9} & \textbf{47.4} & \textbf{63.7} & \textbf{78.1} \\
    \midrule
    \multirow{2}{*}{Model} & \multicolumn{4}{c}{SQuAD-test} \\
     & Top-1 & Top-5 & Top-20 & Top-100 \\
    \midrule
    DPR & 42.5 & 66.8 & 76.2 & 85.0\\
    DCSR & \textbf{49.5} & \textbf{69.6} & \textbf{79.6} & \textbf{86.4} \\
    \bottomrule
    \end{tabular}
    \caption{Performance comparison on both SQuAD-test and SQuAD-dev.}
    \label{tab:squad_dev_result}
\end{table}

By analyzing the training instances, we observe that there exists a severe distribution bias problem in SQuAD: SQuAD-dev and SQuAD-train share a great number of positive passages. In fact, almost all positive passages in the SQuAD-dev could also be found in SQuAD-train. Of all 7921 questions that have at least one positive passage containing the answer in SQuAD-dev, 7624 (96.25\%) of these passages' titles could be found in the positive passages of SQuAD-train. More surprisingly, 6973 (88.03\%) of these passages are shared between SQuAD-train and SQuAD-dev. However, this feature is exactly what SQuAD-test does not have, resulting in relatively poor performance. But again, this phenomenon reveals another strength of our DCSR, that it enjoys better generalization ability than DPR, thus is more robust in practical use.

\section{Case Study} \label{case_study}

To analyze the retrieval performance difference between DPR and DCSR, we especially focus on the different Top 1 predictions on SQuAD. We count the number of winning times for each baseline, where DCSR significantly outperforms DPR (893 vs. 161), shown in Figure \ref{tab:case_study}.

\subsection{DCSR winning cases} On the question \textit{Who was the NFL Commissioner in early 2012?}, the strengths of our DCSR are listed as follows.

$\bullet$ \textbf{Capability of utilizing contextual information.} The key phrase \textit{2012} and \textit{NFL} is faraway from \textit{Commisioner Roger Goodell}, while our DCSR is still capable of capturing such distant contextual information.

$\bullet$ \textbf{Locating the exact sentence of the answer.} This is an obvious feature of DCSR, as we are modeling on the granularity of contextual sentences.

On the contrary, due to \textit{Contrastive Conflicts}, the question encoder of DPR is severely affected that it cannot generate fine-grained question representation. Therefore, on this question, DPR can only find out one key phrase \textit{commissioner}, falling into a totally wrong prediction.

\subsection{DCSR losing cases} On the question \textit{Super Bowl 50 decided the NFL champion for what season?}, our DCSR has already found a contextual sentence that is very close to the given question, with several key phrases detected. However, this contextual sentence is actually a low-quality index, as it suddenly reaches the end of the passage. This is caused by the brute force segmentation strategy of OpenQA, which focuses on the passage level and restricts the length of each passage to 100. In this paper, we perform sentence split directly on these broken passages, which as a result breaks down many sentences into low-quality indexes, affecting the final retrieval performance. We do not intend to refine the splition strategy to have a fair comparison with DPR, and leave it for future investigation.

\end{document}